\documentclass[openany,reprint,NumberedRefs,breaklinks]{JASAnew_arxivmod}

\usepackage{float}  

\usepackage{tabularx}
\newcommand\setrow[1]{\gdef\rowmac{#1}#1\ignorespaces}
\newcommand\clearrow{\global\let\rowmac\relax}
\clearrow

\usepackage{etoolbox} 
\usepackage{breqn}   

\makeatletter
\let\cat@comma@active\@empty
\makeatother

\def\ie{{\em i.e.,} }
\def\eg{{\em e.g.,} }

\newcommand\fakelarge{{FakeLarge}}
\newcommand\fakesmall{{FakeSmall}}
\newcommand\cglarge{{CGLarge}}
\newcommand\cgsmall{{CGSmall}}
\newcommand\realdata{{Real}}

\begin{document}

\title[ConvNets for Counting Transient Phenomena in Steelpans]{ConvNets for Counting: Object Detection of Transient Phenomena in Steelpan Drums}
\author{Scott H. Hawley}
\email{scott.hawley@belmont.edu}
\affiliation{Department of Chemistry \& Physics,  Belmont University, Nashville, TN 37212, USA}

\author{Andrew C. Morrison}
\affiliation{Natural Science Department, Joliet Junior College, Joliet, IL 60431, USA}

\preprint{Hawley \& Morrison, JASA}		

\date{\today}

\begin{abstract}
We train an object detector built from convolutional neural networks to count interference fringes in elliptical antinode regions in frames of high-speed video recordings of transient oscillations in Caribbean steelpan drums illuminated by electronic speckle pattern interferometry (ESPI). The annotations provided by our model aim to contribute to the understanding of time-dependent behavior in such drums by tracking the development of sympathetic vibration modes. The system is trained on a dataset of crowdsourced human-annotated images obtained from the Zooniverse Steelpan Vibrations Project. Due to the small number of human-annotated images
and the ambiguity of the annotation task, we also evaluate the model on a large corpus of synthetic images whose properties have been matched to the real images by style transfer using a Generative Adversarial Network. Applying the model to thousands of unlabeled video frames, we measure oscillations consistent with audio recordings of these drum strikes. One unanticipated result is that sympathetic oscillations of higher-octave notes significantly precede the rise in sound intensity of the corresponding second harmonic tones; the mechanism responsible for this remains unidentified. This paper primarily concerns the development of the predictive model; further exploration of the steelpan images and deeper physical insights await its further application.
\end{abstract}


\maketitle

\section{\label{sec:1} Introduction}
{Electronic Speckle Pattern Interferometry (ESPI) has proven to be an effective technique for musical acoustics research. ESPI provides a means for the measurement and visualization of vibrating plates and membranes making up musical instruments such as violins, guitars, drums, and others.}\citep{Moore2018,Bakarezos2019} ESPI offers the capability of making amplitude measurements for small vibrations; time-averaged ESPI produces images with light and dark fringes which are lines of constant surface deformation proportional to the wavelength of the laser light. These images are similar to Chladni patterns in that they reveal the mode shapes of vibrating surfaces,
(although typically Chladni patterns are used to reveal standing
wave patterns whereas the images in the present paper are of transient phenomena). While lacking the full spatial resolution of traditional film-based laser holography images,\citep{richardson} the relatively low cost and ease of setup make ESPI a popular choice for researchers and educator{s}.\citep{moore_simple_2012}

The use of high-speed video of ESPI images has been applied to the case of Caribbean steelpan drums.\citep{morrison_high_2011} The steelpan drum is a membranophone that originated in Trinidad and Tobago as instrument-makers re-purposed steel oil drums,\citep{aho} stretching the steel into a concave surface and dividing it into a set of flattened, tuned subdomains often referred to simply as ``notes.''  It is played using straight sticks tipped with rubber. When a particular note is struck, waves emanate from the point of impact.  At the boundary for the note, some of the wave energy is reflected and sets up standing waves,\citep{morrison_searching_2017} while the remainder propagates throughout the full steelpan domain and triggers sympathetic vibrations
among the other notes. An accurate characterization of the sympathetic vibration time evolution has yet to be realized.\citep{maloney_2010,monteil_2013}
A fundamental question is how much of the sound of the drum is due to nontrivial time-dependent behavior of
the drum notes (as opposed to steady-state resonant modes).

To better understand the full dynamics at work in the steelpan, high-speed ESPI
images merit closer, quantitative measurements, and yet the enormous quantity of
frames recorded poses a burden on researchers to properly annotate and catalog
what is seen in the images. Thus the ``Steelpan Vibrations Project"
(SVP)\citep{SVP} was formed in partnership with the
Zooniverse.org\citep{borne_zooniverse_2011} platform for crowdsourced data
analysis. Zooniverse arose in the context of large-scale sky surveys of
galaxies, relying on human volunteers from around the world to use a World Wide
Web interface to annotate the images and classify the galaxies seen in the
images.\citep{GalaxyZoo} The specific nature of the annotation used in the SVP
will be described in Section~\ref{subsec:svp}.
As the SVP progressed, it became apparent that an insufficient number of
volunteers were contributing to the project, such that progress in annotating
the large dataset of images was slow.  In addition, because of the variation in
human annotators' work, having multiple volunteers' annotations of the same
image was deemed necessary,\citep{garcia_evaluating_2017} further slowing the
progress of using these annotations to understand the dynamics of the steelpan.
Thus the use of automated annotation methods merited exploration.

While traditional methods of ellipse detection such as
the Elliptical Hough Transform \citep{ell_hough} can
be effective for smooth, well-defined ellipse features,
the noisy and highly variable nature of the ellipse
regions in SVP images, combined with the additional task
{of} counting the rings per antinode, make the Elliptical
Hough Transform a poor fit for this task. There are adaptions
to account for incomplete shapes and noise \cite{rand_hough, rand_ell_hough2},
however the presence of labels via the SVP made us interested in a
machine learning approach. Thus
we sought to adapt methods of neural network based
object detection models to our unique use case.

The success of machine-learning systems at extending the image-annotation efforts of humans
has been demonstrated in a variety of domains.  Notably, image-classification challenges
involving the recognition of handwritten numerical digits\citep{lecun_mnist_2010} and images of various animals and vehicles.
\citep{krizhevsky_imagenet_2012,resnet}
The task of localizing and classifying {\em portions} of images
is known as ``object detection;''\citep{viola_rapid_2001}
typical uses include surveillance systems and satellite imagery analysis\citep{objdet_satellite}
as well as astronomy applications\citep{objdet_astro} such as galaxy classification.\citep{objdet_galaxies}

Multiple algorithms exist for object detection,
and among the most popular and successful in recent years \citep{YOLO,r_cnn,ssd} are those which rely on
convolutional neural networks (CNN) that reduce each
image into a (large) set of learned features that are then fed into a fully-connected layer to predict locations of objects and their classifications.
The scheme used for SPNet is inspired by that of YOLOv2,\citep{YOLO9000} but uses one of
a variety of ``stock" CNN base models, along with a
few important modifications
specific to the domain of ESPI imagery of steelpan drums, and the annotation task of the SVP, as follows: Most object detectors operate on color images of
everyday objects, animals, and people found in datasets
such as ImageNet\citep{imagenet}, whereas the SVP task required the resolution
of constantly-changing patterns in grainy, grayscale images.
Most object detectors provide classifications of their objects,
whereas the SVP task required regression to ``count'' interference fringes. While CNNs are known to perform well at detecting and classifying textures\citep{textures_2006,imagenet_textures} or for
counting numbers of objects or people\citep{people_counting}, their use
to ``count'' rings (or, phrased more carefully, to discover correlations between image patterns and ring counts) which
may have similar ``texture'' but different spatial extents,
was not an application that we observed to have received widespread attention.
Most object detectors make location predictions
for rectangular regions of images, whereas the SVP
required tracking antinodes within elliptical regions. When we began work in 2017,
elliptical object detectors were not in widespread use, however
while preparing this paper a classifier for wood knots was
published\citep{wood_ellipses} which uses a different scheme
from what we present here.

The paper is organized as follows: Section~\ref{sec:spnet_design} presents
details of the SPNet algorithm and training.
Section~\ref{sec:perf_met} presents some performance metrics,
Section~\ref{sec:physics} presents preliminary physics results,
and Section~\ref{sec:disc} provides a discussion of these results.
A separate paper
discussing these and further physics results is in preparation.

For the purpose of reproducibility,
the SPNet computer code is available at \url{https://github.com/drscotthawley/spnet},
and two of the datasets used have been released
on Zenodo.\cite{my_dataset}

\section{\label{sec:spnet_design} SPNet Design}

\subsection{\label{subsec:svp} The Steelpan Vibrations Project (SVP)}

\begin{figure}[b]
\vspace{0.3cm}
\includegraphics[width=\reprintcolumnwidth]{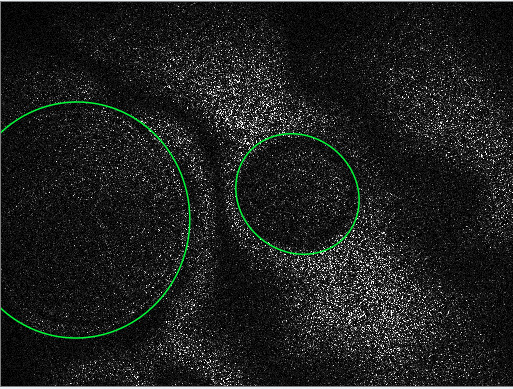}
\caption{\label{fig:1}{(color online) Illustration of Steelpan Vibrations Project\citep{SVP} (SVP) task:
Ellipses ``drawn'' (in green) by human
annotators around antinodes in an ESPI steelpan video frame
via the Zooniverse crowd-sourcing data annotation interface.  {\em Not shown}: Annotations also
include users' counts of the number of interference fringes or rings for each antinode region.}}
\vspace{-.4cm}
\end{figure}

Volunteers recruited for the SVP are presented with randomly-selected frames from
high-speed videos such as the grayscale image shown in Figure~\ref{fig:1}a, and are tasked
with using a web interface to place elliptical boundaries around the antinode
regions (as shown in green), along with counting the number of
interference fringes or ``rings'' for each antinode.
Multiple videos for different steelpan-strikes are available,
which show different regions of the (same) steelpan being excited.

The frames that are included in the SVP are taken from an ESPI optical arrangement and were captured by a
high-speed camera and processed by image subtraction of a reference frame from individual video frames after the drum has been struck. The drum was struck on the back side of the note such that the front side of the note would be unobstructed to the camera's view. The drum was struck with a metal ball driver held by hand at amplitudes well below the typical playing conditions. The vibration amplitudes must be small to be able to be seen clearly in the ESPI frames.

\begin{figure}[b]
\vspace{0.3cm}
\includegraphics[width=\reprintcolumnwidth]{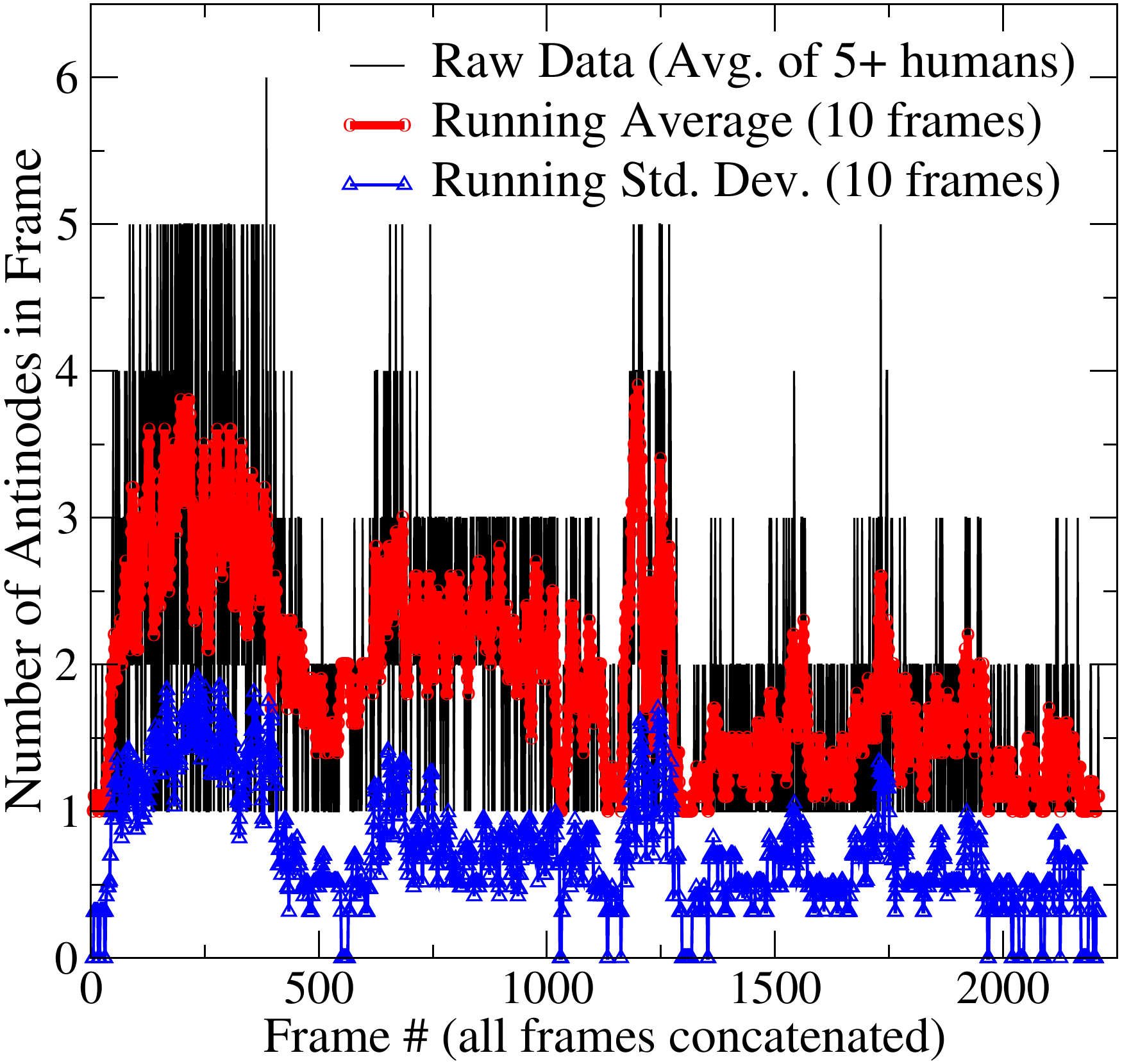}
\caption{\label{fig:an_vs_frame}{{(color online) Graphical representation of one aspect of the variability in the aggregated human annotations comprising the SVP dataset. While, physically, antinodes typically persist over 50 to hundreds of frames, the fine structure of the raw data in this graph shows that the presence of some antinodes may or may not have been annotated consistently frame-by-frame (even in the aggregated data). This is the dataset used to train {\em and score} the SPNet model. This does not display (the further) variability in ring counts, only whether an antinode is marked.}}}
\vspace{-.4cm}
\end{figure}

{For the SVP classification task, organizers required
that at least
15 people supply annotations for a given video frame (image) before it could be analysed for clustering\citep{garcia_evaluating_2017} and then for each
antinode in a frame, at least 5 annotations would be needed. For example, for an image with 3 antinodes, ideally there would be 45 annotations, which were grouped via cluster in X and Y directions. If a volunteer's suggestion was too far from the average (e.g., their mouse slipped) then it was not considered.
Then averages were performed over the ellipse parameters and number of fringes, these averages were written to a file, which
comprised the "raw" or "ground truth" data for training
the SPNet model.
As indicated in Figure \ref{fig:an_vs_frame}, from frame-to-frame, some antinode regions will appear or disappear.  Beyond
variability among volunteers, it can be very much a
``human judgement call''` as to whether
a given ring-shape should be marked as an antinode or not; volunteers were exposed to one frame at a time rather than viewing video.  Even with the benefit of viewing multiple frames, the authors of this paper (who may be considered to provide an ``above-average" level of consistency as annotators), it is not always clear -- especially immediately after a strike -- which shapes to mark as antinodes.  Furthermore, often the struck note would exhibit a ``twin aninode'' structure resulting from its excited 2nd harmonic, in which case annotators may have drawn an ellipse around the whole note, or drawn two ellipses around the two (alternating) sections of the note.
{This is perhaps at variance with the experience of many machine learning enthusiasts who are accustomed to working with very clean datasets and/or well-defined tasks. To better clarify the difficulty of performing the task with consistency, we invite readers to visit the SVP website \citep{SVP} and try annotating several images themselves.}
Continuing our example from above, if 11 of the 15 people missed one of the 3 antinodes, then it would be rejected and not included in the dataset at all for that frame. }

{Regarding the variability in volunteers' ring counts: When we compute the standard deviations of volunteers' ring counts of each antinode and average over all antinodes, we find a value of 1.7. This is considerably wider than the $\pm 0.5$ used for scoring the SPNet model's accuracy, below.  For a standard deviation of 1.7, the area under a normal probability distribution within $\pm 0.5$ of the mean is approximately 0.23, which implies a typical volunteer's ring-count accuracy metric for comparison with SPNet would be 23\%.}

The  task of SPNet is to match (average) human performance from the SVP for the frames
available, as well as to ``fill in'' the missing
annotations for frames in-between those already annotated by volunteers.  The
specificity of this goal will affect the design of the training, discussed
in Section~\ref{subsec:training}
--- the design goal of ``filling in'' missing frames means that the trained SPNet model is not
intended to serve as a generic ``deployable'' inference model for general ESPI images that differ
qualitatively from those in the SVP dataset.  Questions regarding the ability of the
SPNet model to generalize to other ESPI images such as those of
guitars are addressed in Section~\ref{sec:disc}.

\subsection{\label{subsec:spnet_arch} Model architecture}

\begin{figure*}[ht!]
\includegraphics[width=\textwidth]{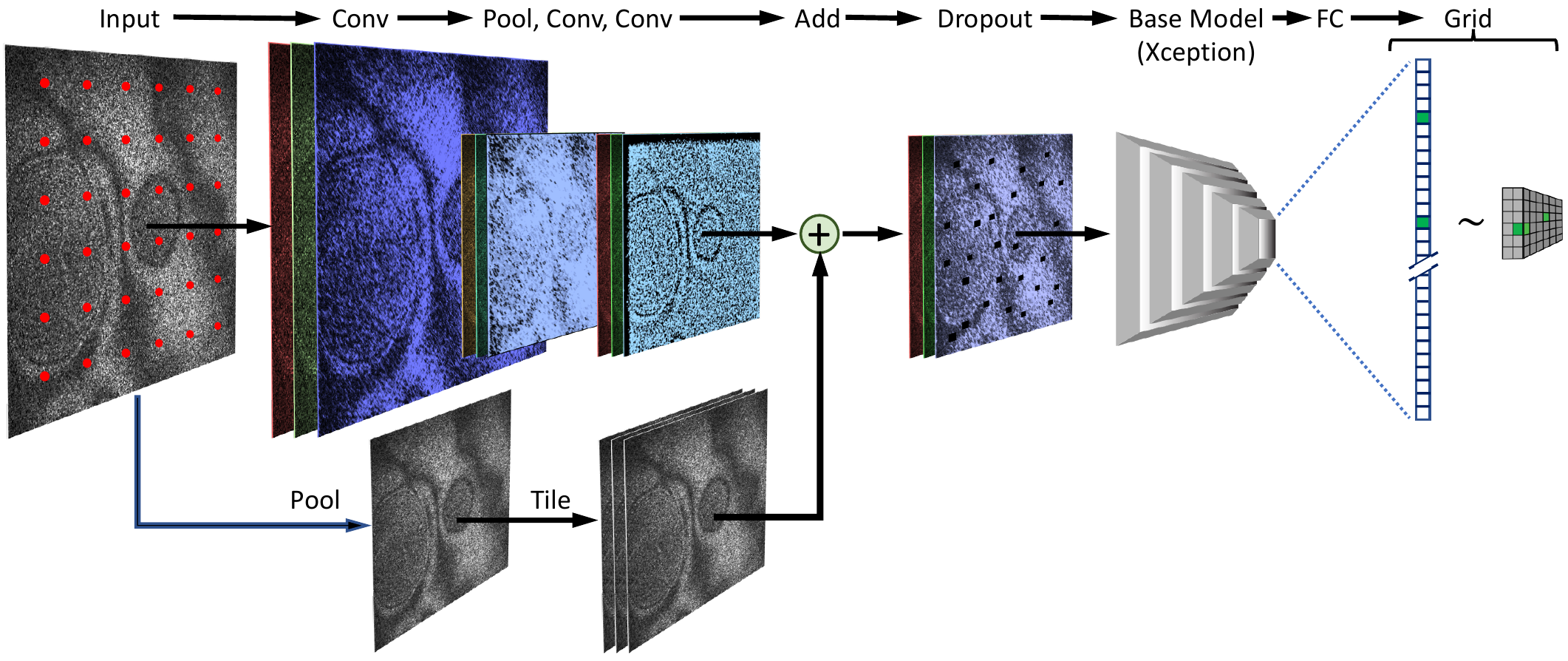}
\vspace{-.2cm}
\caption{\label{fig:spnet_arch}(color online) Diagram of the SPNet architecture.
The grayscale input image is resized via average pooling and two additional (``color") channels
are added via 3x3 convolutions before feeding into a ``stock" base model
chosen from available
Keras models (as described in the text, we prefer Xception\citep{xception}), which is then fully connected
to a flattened layer which holds the values of a 6x6x2 grid of
predictors for the 8 variables in Table \ref{table:vars}. ($6\times 6\times 2\times 8 = 576$ values in the model output.) The operations to the left of the
base model can be regarded as a ``residual block" designed to shrink
the image to lower memory costs while still retaining some
finer details of the larger input image. Also shown as an
array of red dots on
the input image are the centroids of regions covered by the predictors,
which predict antinode centroid coordinates in terms of offsets from these locations. Not shown: Leaky ReLU activations and
batch normalization between layers. (Note: the images
shown for intermediate layers are ``artwork,'' not
actual layer activations.)
}
\vspace{-.2cm}
\end{figure*}

The overall strategy of SPNet is inspired by YOLOv2\citep{YOLO9000}, but the model
differs in
that we use one of several pre-defined 'stock' multi-layered CNN architectures for
the main convolutional network, such as  MobileNet\citep{howard_mobilenets_2017}, InceptionResnetV2\citep{inceptionresnetv2} or 
Xception.\citep{xception}
The ability to easily swap in various predefined CNN base models is made possible via the Keras neural network framework.\citep{keras} These
models can be initialized using random weights or weights pre-trained on
Imagenet.\citep{imagenet_cvpr09}  Our experience indicates that the
Xception \cite{xception} provides a good base model yielding high
accuracy, stable training, and reasonable execution time. (MobileNet, although faster, was not as accurate, whereas InceptionResNetV2 proved both slower and more difficult to train consistently.)
These base models typically expect square-shaped image inputs with 3 color
channels, and large input images can result in networks with so many tunable
parameters (weights) that their memory requirements exceed the capacities
of single computer workstations.
In order to supply input images compatible with available pre-trained
base model architectures while keeping memory requirements manageable,
we first resize our 512x384 grayscale input images to
a square size of 331x331. Even this proves to be unnecessarily and prohibitively
memory-intensive, so we shrink this by a factor of two using ``average pooling," and then
``tile" (i.e., repeat or broadcast) the grayscale channel to form 3
identical ``color" channels -- this is the
lower path shown in Figure~\ref{fig:spnet_arch}.  Doing this alone, however,
could result in some loss in fine detail, so we combine the lower path with the result
of the ``upper path" consisting of multiple 3x3 convolutions yielding 3 filter channels,
in concert with a pooling operation for size reduction.
Adding these two paths forms a ``residual block" \citep{resnet} for which
the lower (pool-tile) path is a skip connection. The skip connection
allows the model to train faster than without it by smoothing the hypersurface of the
loss function \citep{loss_landscape}, and the upper
(conv-pool-conv-conv) path
allows the model to better resolve fine features from the larger (331x331) image
before reducing it in size to feed into the base model.
The pre-processing layers (before the base model)
include Leaky ReLU activations and batch normalization.  We also add a small amount (0.1) of dropout \citep{dropout} before the
base model to help avoid overfitting.

The output of the base model is fully connected to a ``flattened" layer whose elements are taken to represent a ``grid" of
outputs we refer to as ``predictors'' which predict attributes of
relevant antinodes for each subdomain of the image covered by the predictor.
Each predictor predicts 8 values shown in Table \ref{table:vars}: $(p, x, y, a, b, s, c, r)$, where these values are
defined relative to the subdomain associated with each predictor, \ie within each ``grid cell,''
according to Table \ref{table:vars}.
The ``existence" variable $p$ $\in [0..1]$ measures the distinction between the background and an object. The values of $x, y, a$ and $b$ are normalized relative
to the size of the image, and $x$ and $y$ are offsets from the center of each respective grid point.
Instead of the ellipse rotation angle $\theta$, we use the two variables
$c\equiv\cos(2\theta)$ and $s\equiv\sin(2\theta)$ which have the dual advantages of
avoiding any coordinate discontinuity at $\theta=0$ as well as ensuring
{\em uniqueness} given
the $180^{\circ}$ rotational symmetry of the ellipses.\citep{saxena_learning_2009}
These variables are later used in training by optimizing the loss function, which
appears in Section~\ref{subsubsec:loss} as Equation (\ref{eq:loss}).

\begin{table}[ht!]
\begin{tabular}{rcp{0.85\columnwidth}}
\hline
$p$&:& the probability of an antinode's existence within the grid cell, $p\in [0..1]$\\
$x,y$&:& coordinates of the {\it offset} of the antinode's centroid relative to the grid cell's center on the image \\
$a,b$&:& the ellipse's semimajor and semiminor axes $(a >= b)$ \\
$c,s$&:&  $c\equiv\cos(2\theta)$ and $s\equiv\sin(2\theta)$, where $\theta$ is the ellipse orientation angle \\
$r$&:& the number of rings (\ie interference fringes)\\
\hline
\end{tabular}
\caption{\label{table:vars} Definitions of predicted variables.}
\vspace{-0.7cm}
\end{table}

\subsection{\label{subsec:datasets} Datasets}

\begin{figure}[ht]
\centering
\includegraphics[width=.49\textwidth]{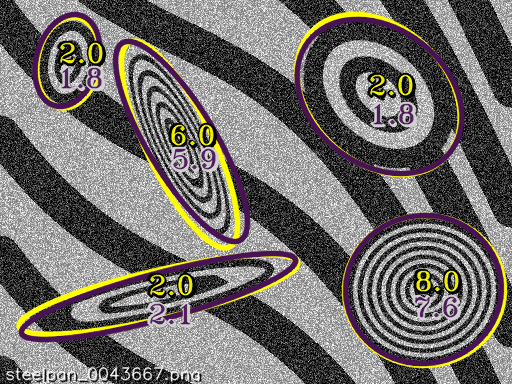}\hfil\break
\includegraphics[width=.49\textwidth]{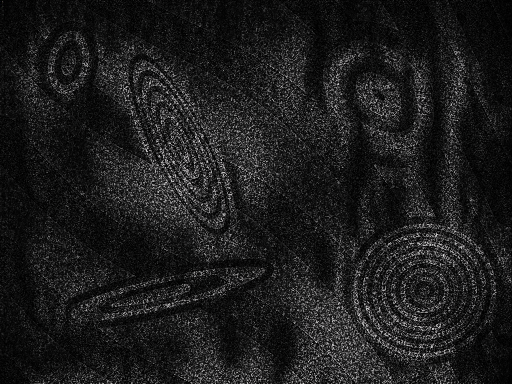}\hfil\break
\caption{\label{fig:synth_data}(color online) Sample fake images, showing ground-truth bounding ellipses and ring counts (upper values, light-yellow) and those predicted by the network (lower values, dark-purple).  Top: original style of fake image, from  \fakelarge{} dataset. Bottom: same fake image with ``real'' style transferred via CycleGAN\cite{CycleGAN2017}, from \cglarge{} dataset.
}
\end{figure}

{Table \ref{table:datasets}
summarizes the datasets uses in this study.}
Early in this project, there were insufficient numbers of
(aggregated) volunteer-annotated images, so in order
to develop and test the model, we procedurally
generated a large (50,000-image) corpus of random ``fake'' images
which combine these salient features: groups of elliptical
rings of varying sizes, orientations, eccentricities,
on a background of wavy patterns, with noise.  (We prefer
the word ``fake'' over the more formal ``synthetic''
to avoid any confusion--
these images are akin to ``artwork'' and have no physical
basis).  Shown in the upper pane of Figure~\ref{fig:synth_data}
is an example of the fake data comprising the \fakelarge{} dataset, along
with superimposed ``exact" annotations (upper values,
light-yellow) and SPNet predictions (lower values, dark-purple.)
The fake images in \fakelarge{} are quite different from the
 \realdata{} data in that the
former have sharp edges and lack the variations in brightness,
contrast, blurriness, and lost pixels observed in the latter.

{Additional datasets were created after it was observed that the model's performance when training on the \fakelarge{} dataset (\eg{} Figure \ref{fig:train_progress}) was much better than when the \realdata{} dataset was used. Questions about the cause of this discrepancy in performance between datasets included:

\begin{enumerate}
\item Was it because there was more data in \fakelarge{} than \realdata{}?
\item Was it because antinode boundaries and rings were much clearer in \fakelarge{} images than in \realdata{}?
\item Was it because the annotations in \fakelarge{} were exact, whereas the annotations in \realdata{} were highly variable, \ie{} similar-looking images in \realdata{} often had very different annotations (thus ``confusing`` the ML system as it trained, and/or causing it to get low scores on evaluations metrics like ``Accuracy''?
\end{enumerate}

To explore the first question, we created the \fakesmall{} dataset, and found that dataset size was not a major factor.
To answer question 2, we matched
the visual properties of the real data (\ie{} the statistics of the pixels in the images)
while still retaining ``exact" annotations against which
to evaluate the model, by training a CycleGAN\citep{CycleGAN2017}
model to do neural style transfer, applying the style of
real images to those in \fakelarge{}. These results were termed
\cglarge{}, one example of which is shown in the lower pane of Figure~\ref{fig:synth_data}.  We also created a corresponding smaller set termed \cgsmall{}.  The implications of this being that the difference in performance between \cgsmall{} and \realdata{} would provide a measure of the degradation in model performance due to the inconsistency in the volunteers' aggregated annotations in \realdata{}. As you will see below, the difference is significant.
}

\begin{table}[]
\setlength{\tabcolsep}{10pt} 
\renewcommand{\arraystretch}{1.5} 
\begin{tabular}{>{\rowmac}c|>{\rowmac}l<{\clearrow}}
\hline
Label & Description   \\ \hline
\fakelarge{} & Fake, 50,000 images           \\
\setrow{\bfseries\realdata{}} & Real data, $\approx$1200 images       \\
\fakesmall{} & 1200-image subset of \fakelarge{}   \\
\cglarge{} & CycleGAN-processed \fakelarge{}       \\
\setrow{\bfseries}\cgsmall{} & 1200-image subset of \cglarge{} \\
\hline
\end{tabular}
\caption{List of datasets, each divided into Train/Validation/Test subsets as 80\%/10\%/10\% splits. Due to RAM limits, all Train subsets contain ~40,000 images, where smaller initial sets have training subsets ($\sim$960 images) augmented  by a factor of 41 to produce ~40,000 images. (see ``Data augment{atation}")
``Fake'' denotes synthetic images, used as a consistent baseline given the inconsistency of the human-annotated ``real'' images.
Bold for the rows indicates that these are the most similar for judging the effects of variability
in the human labels in \realdata{} (whereas those in \cgsmall{} are ``exact").
Datasets \fakelarge{} and \cglarge{} are available from Zenodo;\citep{my_dataset}, whereas release of the \realdata{} Read dataset is delayed pending a future paper.
}
\label{table:datasets}
\end{table}

\subsubsection{Data preparation\label{subsubsec:data}}
We obtained a set of aggregated data from multiple volunteers' annotation attempts; \cite{garcia_evaluating_2017} although the users' ring counts
were entered as integers, the aggregation process produces decimal ring counts.
The main data preparation work for SPNet lay in taking the aggregated SVP data and setting up the correct
vector of target values $Y$ for all grid cells, for all images, in a way that would be
unambiguous and thereby `easiest' for the system to learn.

First, we initialize all predictors to indicate no existence, \ie $p=0$, and for all
other variables to be set in the middle of their respective (normalized) ranges.
Then for each set of annotations (for each image), also called ``metadata,''
we sort the antinodes by their centroid locations, first vertically and then horizontally,
then compute which grid cell each antinode `belongs' to.  For the first of the two
predictors in that cell, we set $e=1$, compute $x$ and $y$ as the difference between
the antinode's centroid coordinates and the center of the grid cell, divided by the width
of the grid cell to keep the values normalized on $-0.5..0.5$.  It is possible that
the Zooniverse interface allowed for $a<b$ and/or for a given rotation angle $\theta$
that may not be bounded within a $180^{\circ}$ range, so for definiteness we swap $a$ and $b$ for any data
in order to enforce $a>b$, subtracting $90^{\circ}$ in the process. After this we compute
$c=\cos{2\theta}$ and $s=\sin{2\theta}$ to enforce the twofold rotational symmetry
of the ellipses as well as avoid any ambiguities with positive or negative angles, or coordinate
singularities at $\theta=0$.  This process is repeated, with the second predictor in a cell
being used if there has already been an antinode found in a given grid cell.  (For more intricate
patterns of antinodes, more predictors per cell and or a more finely-grained grid of predictors
could be used.  The choice of $6\times 6 \times 2$ was found to be adequate for the SVP data.)
 Having set up all the target or ``true'' output data $Y$ for the grid
of predictors to be trained against, it is possible to begin computing a loss function.
First, however, it is necessary to augment the input data to improve the
generalization performance of the model.

\subsubsection{Data augmentation\label{subsubsec:data_aug}}
Augmenting the Training set is a common regularization technique used during the training of machine
learning systems
to increase the variance of a dataset and thus make a trained model more robust, \ie
to improve its generalization performance when operating on new images.  It is crucial in relatively small
datasets such as the $~\simeq 1200$ images obtained from the SVP.
We perform augmentations at two different stages.

The first stage consists of preprocessing augmentations that (randomly) change both the images {\em and} annotations together --
rotations ($\pm 10^{\circ}$),  translations ($\pm 40$ pixels), and reflections -- as well as some image processing such as noise or blurring.

The second set of augmentations are
performed ``on the fly'' at the start of each training epoch, on all input images from the first stage, and consist of random changes to the images {\em only} without altering the annotations,\footnote{The reason for avoiding rotations and translations while training is that these might cause ellipses to
jump discontinuously from one assigned predictor to another; the code could be made to handle that but it currently does not.} such as blurring,
adding noise and
``cutout" \citep{cutout} (\ie excising multiple rectangular subdomains), or
changes to brightness or contrast.
The on-the-fly augmentations applied once per epoch for 100 epochs to $\approx$40,000 training images from each dataset (after the
first set of augmentations) mean that during training the
model is trained on approximately 4 million different images
for each dataset.

\subsection{\label{subsec:training} Training procedure}

\subsubsection{Loss function\label{subsubsec:loss}}
Training is structured as a supervised regression problem using mean squared error (MSE) loss for all variables, subject to a few caveats as follows.
For compactness, we use the symbol  $\Delta^2_u$ to
 denote the squared error
 for a variable $u \in \{p,x,y,a,b,c,s,r\}$, so \eg $\Delta^2_x \equiv (\hat{x}-x)^2$, with predicted values denoted by ``hats.''
 In this notation, we define the loss function $L_j$ for each grid-based predictor $j$, weighted by the
 the ground truth existence $p$ (= 0 or 1) of an antinode in each region, with constant scaling
 factors $\lambda_u$ (tuned by experience so that the terms in the sum are all comparable in magnitude) to be given by:
\begin{dmath}
L_j =
-\lambda_p \Delta^2_p
  + p \left[
 \lambda_{\rm center} (\Delta^2_x + \Delta^2_y)
+ \lambda_{\rm size} (\Delta^2_a + \Delta^2_b)
+ \lambda_{\rm angle} (a-b)^2 (\Delta^2_c + \Delta^2_s)
+ \lambda_r \Delta^2_r \right]
\label{eq:loss}
\end{dmath}
The total loss $L = (1/N)\sum_{j=1}^N L_j$ is then the mean over all predictors $j$, with  $N = 6\times 6\times 2 = 72$ being the total number of predictors in the output grid.
The term in brackets in Eq. (\ref{eq:loss}) is scaled by the ground truth object existence probability $p$ (= 0 or 1), because without existence all other quantities
have no ground truth values. The use of the squared difference $(a-b)^2$ to scale the contribution
due to the angle reflects the intention that, the more circular an ellipse is, the less its angular orientation should matter.

(Replacing the first term in the loss (\ref{eq:loss}) with a cross-entropy term, i.e.,\hfil\break
$-\lambda_p \left[p\log(\hat{p}) + (1-p)\log(1-\hat{p})\right],$
was found to confer no appreciable improvement to the results.)

We also add an L2 regularization or ``weight decay'' \citep{weightdecay_orig,zhang2018three} with strength 1E-4 to all layers in the Keras model;\citep{keras_l2} we find this regularization to be important for avoiding overfitting.

\subsubsection{Model Initialization\label{subsubsec:init}}
Although it is possible to initialize the base model supplied by
Keras using weights pre-trained on ImageNet, the different
nature of our images (grainy grayscale ESPI rather than
color images of common objects, animals, vehicles, etc.) and our intended output type
(regression rather classification) made these pre-trained
weights of little utility, and no better than random initialization. Thus we train all model layers from random
initial weights.

\section{\label{sec:perf_met} Model Performance and Evaluation}

\begin{figure}[ht!]
\centering
\includegraphics[width=.49\textwidth]{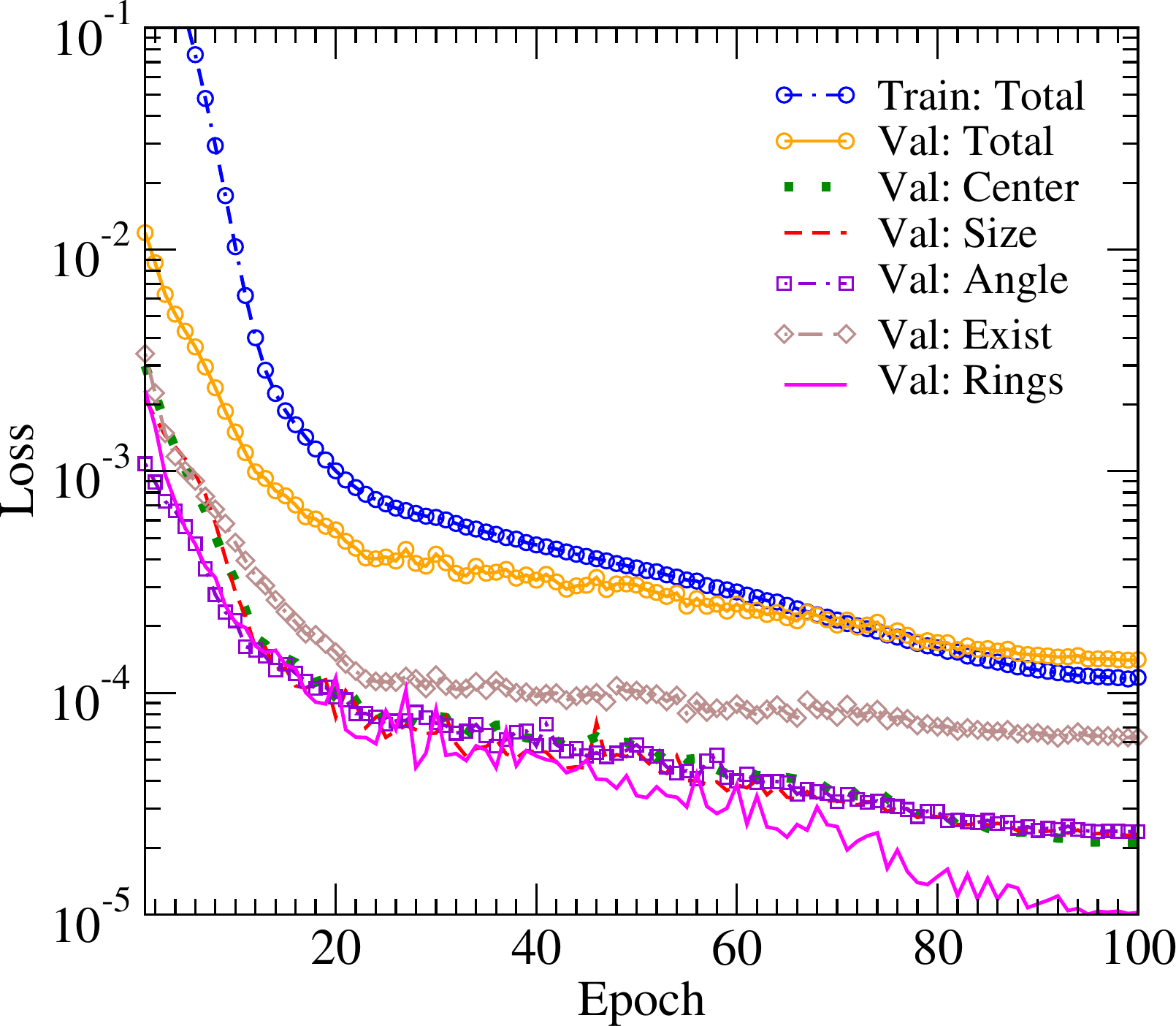}\hfil\break
\hfil\break
\includegraphics[width=.49\textwidth]{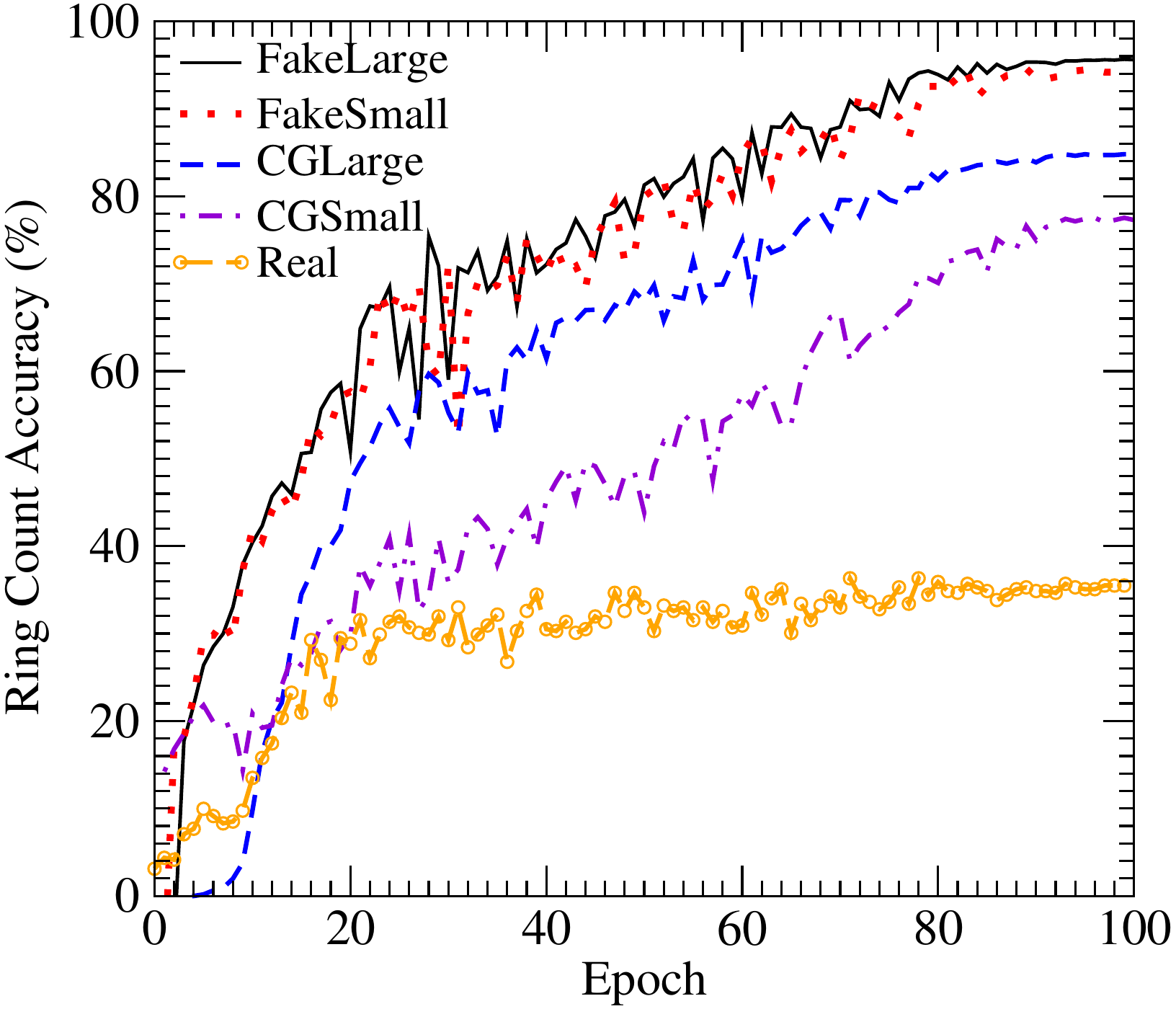}\hfil\break
\vspace{-1cm}
\caption{\label{fig:train_progress}(color online) Training progress. Top: various components of the loss function for dataset \fakelarge{}. (A similar graph for \realdata{} would show Validation loss values leveling off after approximately 20 epochs, which is where the Training loss crosses the Validation loss.)
Bottom: Classification-like accuracy scores for ring counts for validation subsets of all datasets. Despite \fakesmall{}, \cgsmall{}, and \realdata{} all having similar numbers of training images (ca. 1200, when are then augmented as per Section~\ref{subsubsec:data_aug}),
\fakelarge{} and \cgsmall{} have much higher accuracy scores than \realdata{}.
The fact that the accuracy for \realdata{} does not improve beyond Epoch 20 indicates the variability of the human-supplied data annotations.
}
\vspace{-.2cm}
\end{figure}

The purpose of the SPNet model is to help with SVP annotations
with the goal of obtaining physical insight into the motion of drums,
not to lay claim to ``state-of-the-art'' status in object
detection nor win a Kaggle competition, nor to provide a
general utility for generic interference measurements, nor
to offer real-time computational efficiency. Nevertheless,
it is important for a method such as ours to yield reliable
results in a timely manner, and for this reason we provide
measurements of training progress and accuracy scores.
 Sample graphs for training progress in terms of loss (component) values and accuracies are shown in
Figure~\ref{fig:train_progress}.  We typically trained for 100 epochs using an Adam optimizer and
``1-cycle'' learning rate schedule\citep{one_cycle,gugger} with cosine  annealing,\citep{cosine_annealing} using a maximum learning rate of $4e-5$. These runs would take 8 hours on a machine
fitted with an RTX 2080Ti GPU.

Object detection models are usually evaluated
in terms of classification and localization. Given that our task is one of regression rather than classification, many object detection metrics do not
apply directly.  However, we {\em emulate} the task
of an individual human in the SVP (who provided
integer values for ring counts up to 11---in which each integer could be regarded as a class), by considering
whether the model's prediction is within $\pm 0.5$ of
the ground truth value.
\footnote{{Even though ours is a regression problem, the accuracy metric emulates a classification metric, as object detectors are typically classifiers. Our metric of ``within $\pm 0.5$'' makes more sense than rounding predicted and ground truth ring counts each to the nearest integer, so that for example, by our metric, a prediction of 5.4 rings compared to a ground truth of 5.6 is counted as a ``correct'' (or ``matching'') classification rather than an incorrect one.}}
 Using this, we produce a ``ring count accuracy'' metric, as follows: We take the number of matching ring counts between ground truth and predictions and divide it by the total number of ground truth objects (antinodes) in the Validation dataset.  For example, 168 matching ring counts out of 482 ground truth objects would yield an accuracy score of 35\%. For comparison, we noted in Section \ref{subsec:svp} that the standard deviation of individual human volunteers contributing to the aggregated ground truth data imply that a typical volunteer subjected to a similar metric would score a ring count accuracy of 23\%.

For an additional metric that applies only to antinode object detection and not ring counts, we compare the aggregated responses by human users in the SVP to the model predictions, according to the following metrics:
{\em precision} (\ie number of true object detections divided by the total number of objects predicted by the model), {\em recall} (\ie true object detections divided by the total number of objects in the aggregated human data), and {\em intersection-over-union} score (IoU, (\ie the fraction of area overlap between predicted ellipses and their ground truth counterparts).  These can be combined into
a single metric known as the mean average precision (mAP), \citep{bailey_map} which has been averaged over a set of multiple detection thresholds
(\ie{} this mAP is comparable to the ``COCO mAP''\citep{mscoco} with the single category of foreground-vs-background detection).
 These scores are listed in Table~\ref{table:scores}.
 As a baseline comparison, the Real dataset was converted to rectangular bounding boxes and processed using the
 object-detection package IceVision\citep{icevision2020}, yielding similar mAP scores of 0.62 and 0.63 using
 IceVision's Resnet50 and YoloV5 models, respectively.\footnote{{A Google Colab notebook for the baseline
 IceVision-based mAP calculation is available at \url{https://tinyurl.com/spnet-icevision}.} }

\begin{table}[ht!]
\label{table:scores}
\begingroup
\setlength{\tabcolsep}{10pt} 
\renewcommand{\arraystretch}{1.5} 
\begin{tabular}{>{\rowmac}c|>{\rowmac}c|>{\rowmac}c<{\clearrow}}
\hline
Dataset & Accuracy   &  mAP \\
\hline
\fakelarge{}                  & 0.95  &  0.97    \\
\setrow{\bfseries\realdata{}} & 0.35  &  0.67    \\
\fakesmall{}                  & 0.94  &  0.95    \\
\cglarge{}                    & 0.89  &  0.89    \\
\setrow{\bfseries}\cgsmall{}  & 0.77  &  0.78    \\

\hline
\end{tabular}
\endgroup
\caption{Scores for accuracy and mean average precision (mAP) for models trained for 100 epochs from the same random initial weights. ``Accuracy'' is defined as number of matching ring counts (within $\pm 0.5$) divided by total ground truth objects, whereas mAP indicates antinode detection rate\citep{bailey_map} over a range of detection thresholds and is independent of ring count.}
\end{table}

We attribute the low accuracy on the \realdata{} dataset to the inconsistency of human annotations, rather than the size of the training corpus, because scores for \cgsmall{} (which has a similar number of images with similar features but consistent annotations) are significantly higher. The difference
between results for the two datasets becomes even more
striking when one considers that the \realdata{} data has {\em less}
variability in images compared to \cgsmall{}, because for
the former the antinodes in a given video clip stay in
only a finite number of places, and the nature of ``filling
in'' missing annotations between frames implies that images
randomly allocated among the Training and Validation sets will
contain many near-duplicates -- in other words, for the \realdata{} data,
one might expect artificially high scores due to ``cheating.''
In contrast, in \cgsmall{} the antinodes are distributed randomly
everywhere, thus making it more difficult for the model to memorize their existence, locations, and sizes.
Furthermore, increasing the Training set when scoring against
the Validation set for the \realdata{} dataset, for example by combining
the Training port{ion}s of \cglarge{} and \realdata{}, confers no noticeable
change in the evaluation scores, because again, the evaluation
data for Real is highly variable.
Even after ``data cleaning'' by the authors' manually
editing the annotations for all ~1200 images in \realdata{}, there was no uniform consistency, as the annotation
involves many ``judgment calls'' of whether an antinode is present, and if so, how many rings should be counted.
Future annotation efforts may benefit from using more than
one frame at a time, such as viewing the stack of frames
as a 3D volume and annotating via the kinds of software
used in medical imaging and segmentation.
Greater refinement of the model architecture and hyperparameter tuning would likely
produce increases in the already-high evaluation scores on the synthetic datasets (\fakelarge{} through \cgsmall{)}, however,
the limiting factor of the  variability in the \realdata{} dataset's annotations implies that continued
revision of the model would have little effect
on the metrics for the \realdata{} data, from which we wish to
extract measurements of physical phenomena.
{We anticipate that through additional cleaning of the dataset (i.e., improving the consistency of the annotations), that the accuracy score will rise accordingly.}

\begin{figure}[ht]
\includegraphics[width=\reprintcolumnwidth]{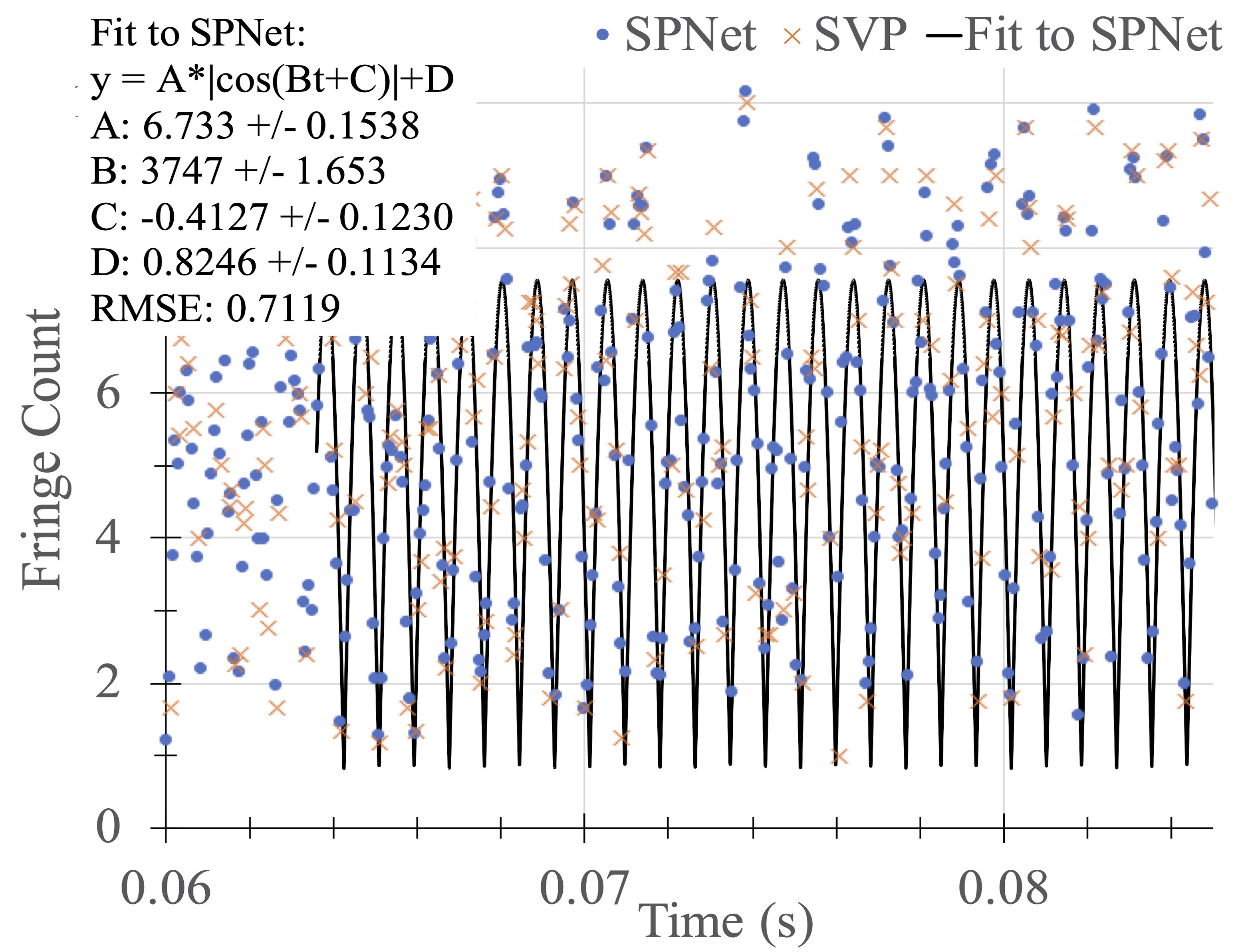}
\caption{\label{fig:fit_inset}{(color online)
A segment of the SPNet amplitude predictions (circles) for the region of the steelpan corresponding to the $D_5$ note is fit with an $|\cos{\omega t}|$ function. The fitting parameter $B$ is equivalent to $\omega$. In this case $f = \frac{\omega}{2\pi}=596$$\,$Hz, which is close to the frequency of the $D_5$ note.}
{(Note that this is a constant-amplitude fit in order to find the frequency.)  Human annotations (SVP, x's) are included for comparison.}
}\end{figure}

Given the difficulty in scoring the model's accuracy
on real data, a concern arises about whether attempts to
extract physics from the model's annotations
are sufficiently warranted.
While this concern merits further study, two
additional consistency
checks give us reasons for optimism.
{Firstly, curve fits of the model's time-series predictions of ring counts for octave notes yield close agreement with the known frequencies of those notes, such as a fit of 660$\,$Hz obtained for the ring counts of the octave note when E$_4\,$($=\,330\,$Hz) is struck, and a fit of 596$\,$Hz for the octave note when D$_4\,$(=$\,294\,$Hz) is struck as shown in Fig.\ref{fig:fit_inset}. The curve fit used was an absolute-value of cosine, which is chosen as the ring count cannot be a negative value. The curve fit successfully matched the frequencies of the octave note for 5 of the 7 recorded strikes for which the SPNet model was used to generate predictions of ring counts. The 2 cases where the model was unable to make predictions leading to a reliable curve fit are due to the amplitude of the drum strike being not sufficiently large enough to generate motion in the octave note that could be detected by the ESPI system.  A graph showing a detail of one such sinusoidal fit is shown in Figure \ref{fig:fit_inset}.}

Secondly, our inspection of the
predictions of the model
when applied to un-annotated video frames
(such as in the sample movie at
\url{https://youtu.be/-rJLwcbQ7Kk}) confirms that SPNet's
predictions of both antinode boundaries and ring counts
are consistent with our own estimations.
Given these reasons for cautious optimism,
in the following section, we begin to explore what physics
may be ascertained when one is willing to take the many
thousands of new frame-annotations provided
by the model at face value, with the caveat that
these are preliminary results.

\section{\label{sec:physics} Preliminary Physics Results}

\begin{figure*}[hbt!]
\centering
\includegraphics[width=\textwidth]{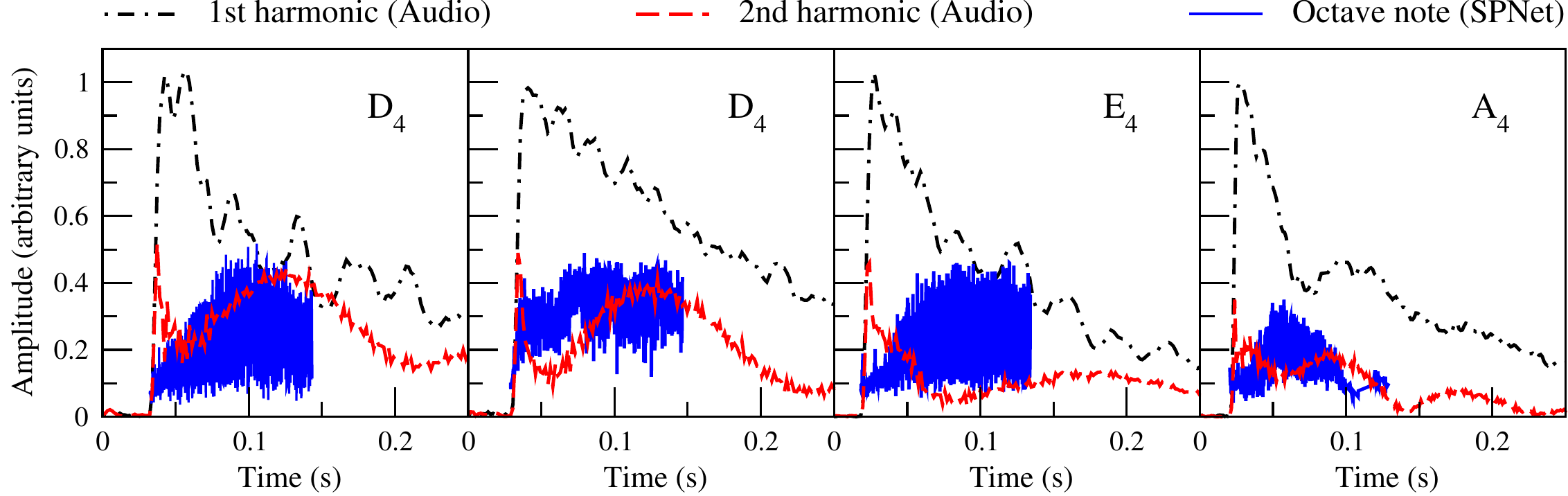}
\vspace{-.2cm}
\caption{\label{fig:amp_vs_time}(color online) Time series for 4 (manual) drum strikes
on fundamental notes D$_4$, D$_4$ again, E$_4$ and A$_4$.
 The solid (blue) line shows the rapid oscillation in ring counts from SPNet's annotations
of the corresponding octave note, for which absolute-value cosine curve fits
show frequencies at or very close to
the expected 2nd harmonic frequencies
({\em i.e.}, D$_5$, E$_5$ and A$_5$, respectively).
Dot-dashed (black) and dashed (red) lines show the amplitude obtained from audio recordings of the events{, for the 1st and 2nd harmonics, respectively}.
The richness of the drum's behavior is evident from the variability between strikes. All graphs show a rapid
damping of the 2nd harmonic immediately
after the initial strike, yet the later rise
in the 2nd harmonic sound intensity significantly lags (or is even uncorrelated with)
the motion of the corresponding octave note
observed in video analysis by SPNet.
Even in the left-most graph where the two appear to correspond,
the lag is significantly longer than would be suggested by
physical delay mechanisms such as wave travel time.
We discuss these further in Section~\ref{sec:disc}.
(The SPNet annotations end before the audio recordings because high-speed video was only recorded for $\approx 150\,$ms.)
}
\end{figure*}

Figure~\ref{fig:amp_vs_time} shows
drum oscillation amplitude as a function of time,
comparing ring counts obtained from SPNet with
audio recordings using a microphone placed 1m from the
center of the drum. Each recording of a drum strike was made using an ACO Pacific Model 7012, $\frac{1}{2}$” condenser
microphone controlled by a custom LabView program triggered to coincide with the high-speed ESPI recording.
Audio recordings were made at a sample rate of 44100 Hz, and analysis of the recordings was done with
the SNDAN package.\citep{Morrison2013a} The large fundamental
note such as that shown in the left of the field of view (as shown in Fig. \ref{fig:1}) is struck, and the SPNet
analysis tracks the rings in a note to the right,
corresponding to the second harmonic.  (This was confirmed by measuring the frequency of the oscillations
in the ring counts.)

\begin{figure}[ht]
\centering
\includegraphics[width=.49\textwidth,trim=0.5cm 0.3cm 0.2cm 0.5cm, clip]{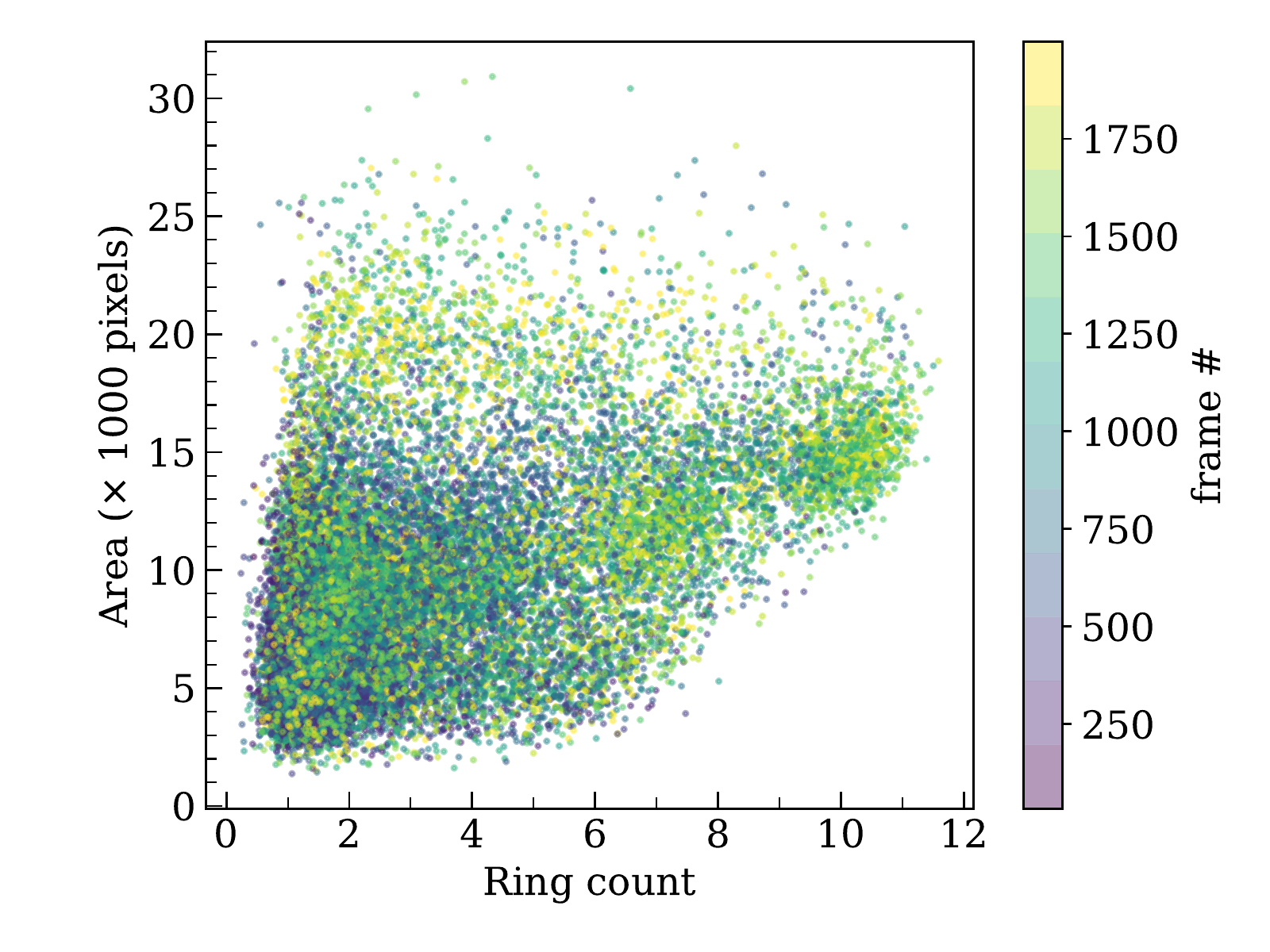}\hfil\break
\vspace{-.7cm}
\caption{\label{fig:area_vs_rings}(color online) Area vs. number of rings
for antinodes detected in four separate videos of drum
strikes.  The largest ring-counts are associated with
large areas, however as expected, the reverse is not the case, for
physical reasons described in the text.  We color the
points by the frame number in each video, as a way
to investigate how the distribution of area vs. ring count
might change over time (although we make no claims for this figure).
}
\end{figure}

Figure~\ref{fig:area_vs_rings} explores the relationship
between the number of rings and the size (area) of the
antinodes. For large ring counts, which indicate
large deformation (or velocity) of the surface, one
would expect the area of the antinode to be the
same as that of the note itself.  Small areas and small
ring counts could result from small notes, or could
result from large note areas in which the note is barely
moving. In the latter case, one would only see the
shape of the largest portion of the note that ``peeks up"
above the threshold set by the laser interference.
It is not obvious, then, what the relationship between
area and ring count should be,  and thus we provide Figure~\ref{fig:area_vs_rings} as a set of raw observations.
The differing coloration of the dots is primarily to allow for articulate viewing ({\em i.e.}, so the reader is not presented
with a large wash of undifferentiated uniform color)
and also to provide the opportunity to observe any time
dependence in the distribution of the values. We do not
claim to detect a noticeable time-dependent trend in
the case of this figure, however in the following figure
there does appear to be some noticeable time-dependence.

\begin{figure}[ht]
\includegraphics[width=.49\textwidth,trim=0.5cm 0.3cm 0.2cm 0.5cm, clip]{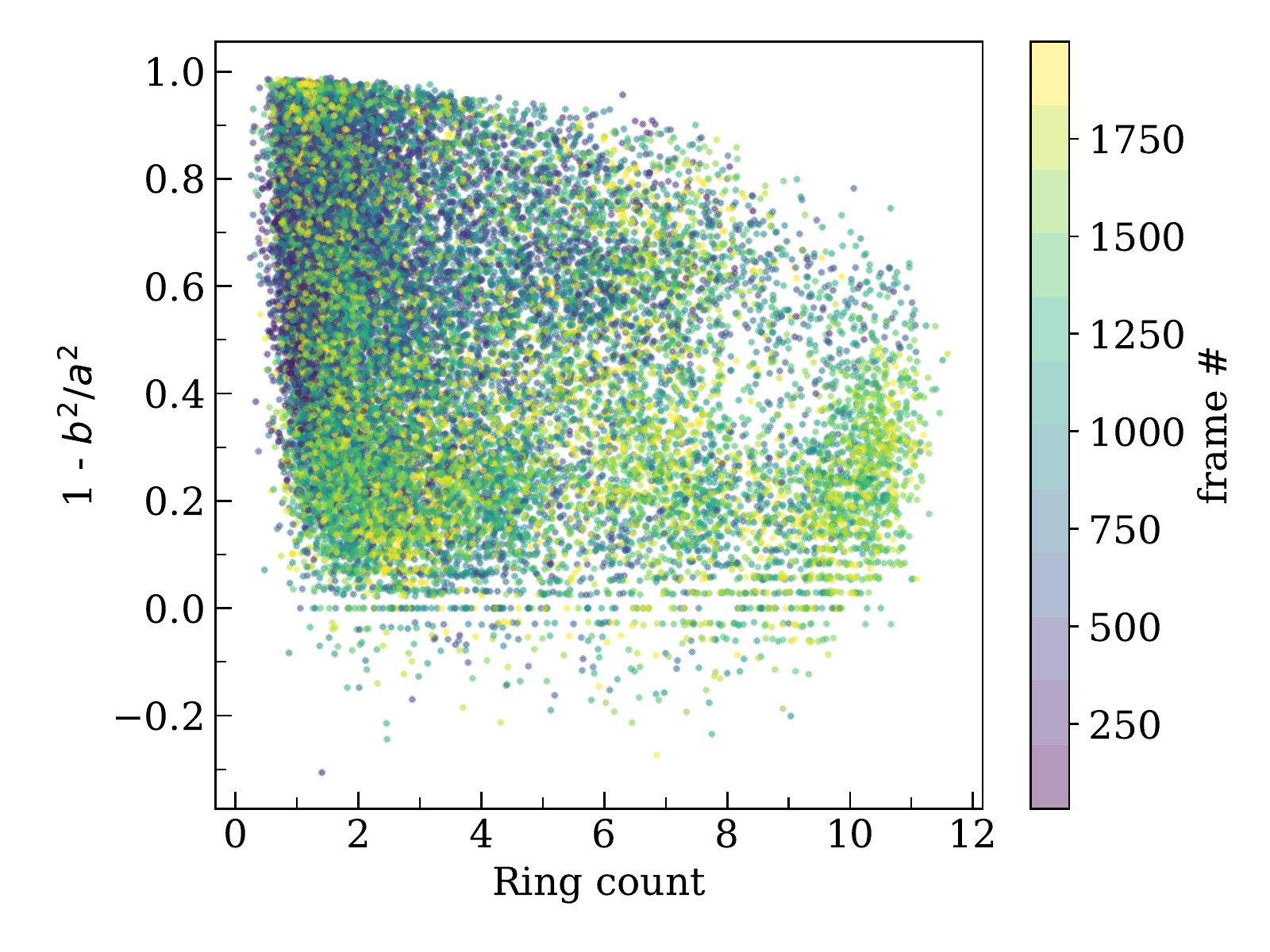}
\caption{\label{fig:ecc_squared_vs_rings}(color online)
Eccentricity squared vs. ring count. We observe that
dark-colored dots representing antinodes at early times
tend to cluster towards the upper left area of higher eccentricity and
low ring count, whereas lower eccentricities with larger ring
counts are seen mostly at later times (lighter colors).
For nearly circular antinodes, the model is not prevented
from predicting $b>a$ sometimes (by as much as 7 pixels),
despite being trained on data for which $a>b$ is always satisfied; hence the negative values of $1-b^2/a^2$.
}
\end{figure}

In Figure~\ref{fig:ecc_squared_vs_rings} we investigate
the relationship between (squared) eccentricity and ring count.
As with area vs.
ring count, it is not obvious what the relationship between
eccentricity and ring count should be: If eccentricity were
 determined purely by the shape of each note,
then we would expect
a ``quantized" set of eccentricities (one for each note-shape),
but instead we see a wide range of antinode eccentricities present.  (The horizontal banding near the
bottom is a non-physical artifact of pixel-integer math.)
In the case of this figure, we observe that the darker
dots representing early times tend to cluster in the upper
right area of high eccentricity and low ring count,
whereas the domain of low eccentricity and high ring count
tends to be occupied only at later times.  We will
discuss this further in Section~\ref{sec:disc}.

A sample movie of SPNet-annotated video frames is available in Supplemental Materials, as well as
at \url{https://youtu.be/-rJLwcbQ7Kk}.

\section{\label{sec:disc}Discussion}

\subsection{\label{subsec:phys_disc}Physical Interpretation}
In Figure~\ref{fig:amp_vs_time} we observe the
differing behavior of the second harmonics when measured
via audio versus SPNet (via the latter's ring-count
annotations of the octave note).
We find this difference surprising, as we would expect these two
signals to exhibit close similarity.

In \ref{fig:amp_vs_time}a, the audio signals for the first and second harmonics initially decay at approximately the same rate ({\em i.e.}, they have the same ``reverberation time''), suggesting that this initial
transient in the second harmonic sound results from the first
harmonic note ringing down as a superposition of first and
second harmonics, and only later does the octave note
in the drum begin to oscillate significantly -- on
differing time scales of roughly 50$\,$ms after the strike for in the video (SPNet output), and 150$\,$ms
for the audio.
The strike shown in Figure~\ref{fig:amp_vs_time}b
exhibits qualitatively different behavior from
the previous graph. The second harmonic in the audio
initially decays much faster than the fundamental,
and rises again ~90$\,$ms later,
whereas the octave note as measured by SPNet begins oscillating immediately and maintains its amplitude.

The drum strikes were performed by hand, not always with
the same velocity or even exactly in the same location
within each fundamental note, so that
differences between Figure~\ref{fig:amp_vs_time}(a) and (b) need not merit consternation in their own right.
One may add to this the fact that these two different
drum notes were hand-hammered by the steelpan tuner and thus there is no
guarantee of consistency from one note to the next.

Allowing for such variations, however, the difference
(for each strike) in the second harmonic between audio
and video (SPNet annotation) measurements is nevertheless noteworthy. At present, we are unable to account for this discrepancy.  Looking at the graphs of the signals (physics aside),
one might propose some kind of ``delay'' of at least 50$\,$ms between the two signals, at which point
it is worth ruling out two mechanisms that would
{\em not} produce such a long-term effect.
Firstly, the travel time of sound in air from the drum to the microphone is no more than 3$\,$ms since the latter is only 1$\,$m away. Secondly, the wave speed
in the drum is roughly 3000$\,$m/s (this is not a precise number because steelpans like the one used in this study are hand-hammered by artisans and thus contain variations in thickness), whereas the distance between the fundamental and octave notes is at most a few centimeters, yielding a wave travel time in the drum on the order of 10$\,\mu$s.
Given that the dynamical timescales for wave travel are so short, it seems unlikely that the difference in signals can
be accounted for in terms of a delay due to wave propagation. Thus an understanding of the physics producing the
observed difference in the measurements for the second harmonic awaits further study.

Turning our attention to the distribution of eccentricity versus ring count
as shown in Figure~\ref{fig:ecc_squared_vs_rings}, we
observe an apparent trend of clustering of early-time antinodes
towards the upper left, with the lower right consisting
of mostly later-time antinodes. This raises several
questions, as the interpretation of this observation
is not straightforward.
While this trend is truly present in the data (and not
some artifact of the order in which points are plotted),
we prefer caution about drawing physical inferences
from this. The idea that large, circular antinodes are
the ones likely to persist the longest seems well-motivated,
but the evolution of a single antinode is not trackable in this
figure: We saw in the second harmonic graphs of
Figure~\ref{fig:amp_vs_time}, ring-counts not only
decrease with time via damping but can increase over time.
(Also, since the frames show oscillating antinodes, each
dot in the graph oscillates left-and-right ``rapidly" in
this figure, regardless of any longer-term trends).
Apart from the ``path" through this graph-space that
an individual antinode might take over time, it is unclear
whether ``missing'' data points have any physical significance.
For example, in this data there is an
absence at early times of low-eccentricity antinodes
with high ring counts, and yet we know that the note
at which the drum is struck oscillates with an essentially
circular shape, with high ring count. Is it then the
case that the ``missing'' circular, high-ring-count
antinodes do not occur, or is it merely that these are
not detected ({\em i.e.}, false negatives) by the
model? The latter scenario seems likely, given that
many volunteers in the SVP failed to annotate the
large initial strike area.  One might similarly
conjecture whether the ``hole" seen around the
coordinates (9, 0.4) is physically interesting,
or is a mere artifact of the available notes on the
drum ({\em i.e.}, the finite number of notes, and/or the
choice of the experimenters on which notes they
recorded), or an artifact of the object detector.
These questions bear further investigation.

\subsection{\label{subsec:ml_disc}Machine Learning Considerations}
Rather than producing a generic object detector package
for measuring interference
fringes in all forms of musical instruments illuminated by ESPI,
we have trained a model to assist in filling in missing annotations
(``in-between frames'')
for a small set of videos focused on a particular region of a
particular steelpan drum.  While the methods used in this paper
could be replicated in other domains if sufficient training data
({\em i.e.}, annotated video frames) were available, the question
of how well our model, trained on such images as we have,
could predict interference fringes in more general situations, remains
open. One would hope that transfer learning\citep{Wang2020Pay} could be applied  using our model as a starting point for similar ESPI images, lowering the requirement for new training data.
Earlier we stated that using transfer learning using ImageNet weights proved no better than starting from scratch,
but the similarity between ESPI images (vs. their difference from typical ImageNet images) could prove beneficial.

Not all instruments exhibit elliptical-shaped antinode regions,
however, we conjecture that the shape is not a primary limiting
factor if one wishes to count fringes apart from requiring
precise bounds on the antinode regions.  Some early work
we performed using image-segmentation model Mask-RCNN\cite{mask_rcnn,matterport_maskrcnn_2017} indicated it could
find peanut-shaped and triangle-shaped antinode regions even
when trained on ellipses, however the code structured on a
deep level as a classifier and we elected not to try to modify it for regression.

The variability in the human annotations of the real data prevented
us from objectively scoring highly when evaluating the model
(because even the testing set exhibited the same inconsistencies),
and although using the fake data (particularly \cgsmall{}) allowed us
to gauge how well the model might perform on consistently-annotated
ESPI images, this fake data was not physically-motivated.
An alternate path to obtaining physically-realistic training data
would be to perform physical simulations of the steelpan
\cite{Kagawa2012} via methods such as Finite Element modeling\cite{gay_2008,french_sim}
and then apply ``styling'' techniques
such as CycleGAN to make the fake images look like the real ones.

\section{\label{sec:concl}Conclusions}

Using an object detector comprised of convolutional
neural networks, it is possible to locate and track antinode
regions on oscillating steelpan drums, and to solve
the regression task of estimating the number of interference
rings in each antinode. While variations in the
human annotations prevented high scores on accuracy
metrics, our ``SPNet" model's performance was sufficient to
extract oscillation information at the correct frequencies
in highly time-dependent, transient regimes.
Data from our analysis indicate a significant
discrepancy between audio recordings of second harmonic
oscillations (sympathetic to a drum struck on a fundamental
note) and optical measurements ({\em i.e.}, video
frame analysis by SPNet).  Explaining this discrepancy
in terms of likely physical processes remains beyond
the scope of our current effort.  Subsequent analysis
published in future papers may reveal additional insights.

\begin{acknowledgments}
The authors thank Thomas R. Moore for helpful consultation during the completion of this paper,
and Matthew Lange who assisted A.C. Morrison with data processing and was supported by NSF grant 1741934.
This publication uses data generated via the Zooniverse.org platform, development
of which is funded by generous support, including a Global Impact Award from Google,
and by a grant from the Alfred P. Sloan Foundation.
\end{acknowledgments}

\bibliography{Manuscript_rev2}

\end{document}